%% file: MLATPSurvey-clean-arxiv.tex
\documentclass{llncs}
\usepackage{amssymb,amsmath}
\usepackage{todonotes}
\usepackage{url}
\usepackage{comment}
\usepackage[colorlinks,urlcolor=blue]{hyperref}
\usepackage{mathabx}
\input{thol.tex}

\usepackage{pgfplots}
\usepackage{tikz}
\pgfplotsset{compat=1.15}

\newenvironment{mypar}[2]
  {\begin{list}{}%
    {\setlength\leftmargin{4mm}
    \setlength\rightmargin{4mm}}
    \item[]}
  {\end{list}}

\title{Learning Guided Automated Reasoning:\\ A Brief Survey}
\author{
Lasse Blaauwbroek\inst{1,2} \and
David Cerna\inst{3} \and
Thibault Gauthier\inst{1} \and
Jan Jakub\r{u}v\inst{1,4} \and
Cezary Kaliszyk\inst{4} \and
Martin Suda\inst{1} \and Josef Urban\inst{1}}
\institute{Czech Technical University in Prague\\
  \and
  Radboud University Nijmegen\\
  \and
  Czech Academy of Sciences Institute for Computer Science\\
  \and
University of Innsbruck
}

\makeatletter
\renewcommand\section{\@startsection{section}{1}{\z@}%
                       {-12\p@ \@plus -4\p@ \@minus -4\p@}%
                       {8\p@ \@plus 4\p@ \@minus 4\p@}%
                       {\normalfont\large\bfseries\boldmath
                        \rightskip=\z@ \@plus 8em\pretolerance=10000 }}
\makeatother

\begin{document}
\maketitle
\begin{abstract}
  Automated theorem provers and formal proof assistants are general reasoning systems
  that are in theory capable of proving arbitrarily hard theorems, 
  thus solving arbitrary problems reducible to mathematics and logical
  reasoning.
In practice, such systems however face large combinatorial explosion, and therefore
  include many heuristics and choice points
that considerably influence their performance. %
This is an opportunity for trained machine learning predictors, which can guide %
the work of %
such reasoning systems.
Conversely, %
deductive search supported by the notion of logically valid proof
allows one to train %
machine learning %
systems on large reasoning %
corpora.
Such bodies of proof are usually correct by construction and when combined with more
and more precise trained guidance they can be boostrapped into very
large corpora, with increasingly long reasoning chains and possibly
novel proof ideas.

In this paper we provide an overview of several automated reasoning and theorem proving domains %
and the learning and AI methods that have been so far developed for them.
These include premise selection, proof guidance in several settings, AI systems and feedback loops iterating between reasoning and learning,
and symbolic classification problems.
 \end{abstract}

 \section{Introduction}

\begin{mypar}{8mm}{5mm}
{\it
  No one shall drive us from the semantic AI paradise of computer understandable math and science!
  
{\hfill\rm--  AGI'18~\cite{UrbanAGI18}}
}
\end{mypar}

Automated Reasoning (AR) \cite{DBLP:books/el/RobinsonV01} and
Automated Theorem Proving (ATP) systems are general AI systems that
are in principle capable of solving arbitrary mathematical and
reasoning problems. Their theoretical \emph{completeness} means that \emph{any} solvable problem, regardless of its
difficulty, will be eventually solved. %

In practice, today's AR and ATP systems however soon encounter
combinatorial explosion, typically preventing them from solving hard
open problems in a reasonable time. Regardless of which proof calculi
they implement, they have to make many decisions concerning which
theories and lemmas to use, which inference steps or tactics to
choose, which instantiations to apply, how to divide the problems, split
the clauses, propose useful lemmas, etc. These choices do not typically %
influence the
theoretical correctness and completeness of the proof mechanisms but
rather the systems' efficiency and practical performance. %

AR/ATP practitioners have often \emph{manually designed and optimized}
a spectrum of heuristics, efficiency and calculus improvements based
on their practical and theoretical insights.  Such manual designs can
sometimes be very successful, as witnessed, e.g., by the large
performance boosts brought by using the right term orderings
\cite{DBLP:conf/icms/JakubuvK18}, literal selection function
\cite{DBLP:conf/cade/HoderR0V16} and age-weight ratios
\cite{DBLP:conf/cade/McCune90} in ATPs, sophisticated indexing methods
for ATPs \cite{DBLP:books/el/RV01/RamakrishnanSV01} and SAT solvers
\cite{DBLP:conf/dac/MoskewiczMZZM01}, as well as by human-designed
improvements of the underlying calculi such as CDCL
\cite{DBLP:conf/iccad/SilvaS96,DBLP:conf/aaai/BayardoS97} in SAT
solving, ordering-based constraints
\cite{DBLP:books/el/RV01/NieuwenhuisR01}, and AVATAR-style splitting
\cite{DBLP:conf/cav/Voronkov14,DBLP:conf/cade/RegerSV15} in
saturation-based ATP.

On the other hand, general mathematics is undecidable and arbitrarily
complicated, and it seems increasingly hard to manually design more
complex heuristics for more complex domains and problems. At the same
time, \emph{automated design and optimization} of parameters,
heuristics, functions and algorithms has been a major topic in AI
since its beginnings. Especially in the last decades the field of
machine learning (ML) has produced a number of interesting
\emph{data-driven} \cite{KMB} methods that can be used in AR and
ATP. Perhaps the most interesting area of such research is how to
combine and interleave the AR and ML methods, creating feedback loops
and meta-systems that can continuously improve their %
skills
and keep finding---for long time---harder and harder proofs and explanations.

In this survey, we cover the development of such methods starting from
the early AI/TP systems and tasks such as high-level knowledge
selection, %
to the today's spectrum of architectures and tasks that include
guidance at various levels, a variety of learning-based methods with
various speed/accuracy trade-offs, and a variety of combinations of
the ML and AR methods.  Automated Reasoning is however a large field
and this brief survey does not make claims to be exhaustive. We mainly
focus here on the fields of Automated and Interactive Theorem
Proving. For some related AR fields %
similar surveys and overviews have been written recently. In
particular, we recommend the recent exhaustive
overview~\cite{Holden21} discussing ML methods in the context
of SAT and QSAT solving.

\section{Early History} %
\label{hist}
According to Davis~\cite{davis2001early}, %
in the beginning of
AR and ATP, two research directions emerged: (i) \emph{heuristic}/AI
emulation of human thought processes, such as Newell's and Simon's
Logic Theorist \cite{newell1956logic} and later Bledsoe's work \cite{bledsoe1972computer}, and (ii)
design of crisp algorithms based on logic transformations and calculi,
such as the early proof procedures by Davis and Putnam \cite{davis1960computing}, Gilmore \cite{gilmore1960proof}
and Robinson \cite{robinson1965machine}. The latter (\emph{logicist}) approach largely
prevailed in ATP in the first decades, thanks to major advances such
as resolution, DPLL and paramodulation. Interesting examples of the
heuristic/AI systems at that time include Lenat's Automated
Mathematician (AM) \cite{Len76-Thesis}, Langley's Bacon \cite{Langley1977BACONAP} and Colton's HR \cite{ColtonBW99}, which propose
concepts and conjectures, thus also qualifying as symbolic ML. %

First interesting combinations of state-of-the-art ATPs with ML methods
(both statistical and symbolic) were designed in the 90s by
the Munich AR group \cite{DS1996b,DenzingerFGS99}.  The methods and
systems included, for example, the invention of \textit{tree (recursive) neural
networks (TNNs)} for classifying logical expressions by Goller and
Kuchler \cite{DBLP:conf/esann/Goller99}, and the development of the E
prover \cite{Sch02-AICOMM} by Schulz, which allowed proof guidance by symbolic patterns
abstracted from related proof searches \cite{Sch00}. Related to that was the %
\emph{hints} proof guidance in \textit{Otter} developed for attacking open conjectures by Veroff~\cite{Veroff96}, again based on symbolic abstraction (subsumption) of lemmas in related proofs.

At the same time, large ITP (\textit{interactive theorem proving}) libraries
started to appear, with the \textit{Mizar} project \cite{BancerekBGKMNPU15} producing over 700 formal
mathematical articles by 2001. Such large libraries have since become
a natural target for combining ML and AR. Urban's 1998 MSc
thesis \cite{urban98} was perhaps the first attempt to learn symbolic
heuristics over such formal libraries with the use of \textit{inductive logic
programming} (ILP -- a symbolic ML approach). Already before that, the
ILF project \cite{Dah97-old} started to work on translations between Mizar and
ATPs. This was continued by the MPTP project
\cite{Urban03,Urb04-MPTP0} which in 2003 released a dataset of about 30000 Mizar ATP problems, and reported about the
first large ATP and ML experiments over it. This included training a \textit{naive Bayes} (Section~\ref{nb}) predictor over the Mizar
library (Mizar Proof Advisor) to select suitable library facts (premises) for proving the next Mizar problems.
In 2004, the Isabelle/Sledgehammer was developed \cite{MengP04}, including the first heuristic (non-learning)
premise selection methods (MePo) \cite{MengP09}.
Such \emph{hammer} systems connecting ITPs with ATPs have become an AR topic of its own, see \cite{hammers4qed}. %

Since efficient premise selection is a critical task in hammers, a lot
of the initial AI/TP work focused on it. The 2007 MPTP
Challenge\footnote{\url{https://www.tptp.org/MPTPChallenge/}} followed
by the 2008 CASC LTB (large theory batch) competition introduced 
benchmarks of large-theory problems and suitable settings for AI/TP
system training and evaluation. This quickly led to fast non-learning methods
such as SInE \cite{HoderV11} integrated in ATPs, as well as larger meta-systems such as MaLARea \cite{Urban07,US+08} that
interleave proving with learning of premise selection.

The early non-learning systems typically focus on symbols and symbolic
features in the formulas, with MePo and SInE using metrics such as
symbolic overlap (Jaccard index) and symbol rarity, starting with the
conjecture symbols and recursively adding the most related formulas
until a certain size of the set of premises is reached. The Mizar
Proof Advisor and \textit{MaLARea} instead initially trained naive Bayes to
associate the conjecture symbols (features) with the premise names,
later adding more complicated syntactic features such as term walks
and also semantic model-based features (Section~\ref{sf}), obtained by
evaluating both conjectures and premises in a growing set of finite
models.

These first approaches turned out quite successful, with MePo,
SInE and MaLARea considerably increasing the performance of the
underlying ATPs in large theories. This has (re-)opened several
research topics, such as (i) how to suitably characterize mathematical
formulas and objects, (ii) what are the suitable ML methods, (iii) on
what level should the ML guidance be applied, and (iv) how to
construct larger feedback loops and meta-systems combining ML and
AR. In the next sections we discuss some of these topics.

.

\section{Characterization of Mathematical Knowledge}
How knowledge is represented is often essential to %
understanding and gleaning deeper insights about the subject encompassing it. To take a prominent example, consider Fermat's Last Theorem. The theorem statement concerns what most would classify as number theory, yet the proof lives in the world of \textit{elliptic curves}. There may be an elementary %
number-theoretic  %
proof of the theorem. However, given the state of mathematics at the time, the shortest path was through a, at least to the mathematics dilettante, seemingly unrelated area. When developing %
methods for classifying mathematical %
expressions, analogously, proper representation is essential to extracting the necessary semantic notions. 
\subsection{Syntactic Features}

As mentioned above, the work on premise selection %
started with syntactic characterizations of conjectures and
premises. The early methods extend the extraction of symbols (already
mentioned in the discussion of SInE and MePo in the previous section)
by using subterms as additional features, applying various
normalizations to them to increase the feature matching between the formulas,
and using syntactic
representation of types and their connections. The aim of
such early investigations %
was to provide HOL Light with automation (a hammer) for proof
development in the Flyspeck project. For example,~\cite{KaliszykU14} %
included features based on the subterms %
normalized by replacing variables by their
types, their de Bruijn numbers, or merging all variables.
As an example, the HOL theorem \texttt{DISCRETE\_IMP\_CLOSED} with the HOL Light statement:
\begin{holnb}\holthm{
!s:real^N->bool e.
        &0 < e /\ (!x y. x IN s /\ y IN s /\ norm(y - x) < e ==> y = x)
        ==> closed s
}\end{holnb}
is characterized by the following set of strings that represent unique features.
\begin{holnb}
"real", "num", "fun", "cart", "bool", "vector_sub", "vector_norm",
"real_of_num", "real_lt", "closed", "_0", "NUMERAL", "IN", "=", "&0",
"&0 < Areal", "0", "Areal", "Areal^A", "Areal^A - Areal^A", 
"Areal^A IN Areal^A->bool", " Areal^A->bool", "_0", "closed Areal^A->bool",
"norm (Areal^A - Areal^A)", "norm (Areal^A - Areal^A) < Areal"
\end{holnb}

The above approach resulted in many unique features, for which various
feature weighting schemes were explored for efficient use with the
premise selection predictors. The most efficient schemes were based on
the linguistic TF-IDF~\cite{KaliszykU13a}, raising considerably the
performance of the best $k$-nearest neighbor predictors. In TF-IDF, a
term $t$, present in a collection of documents $D$, is weighted by the
logarithm of the inverse of the term's frequency within $D$, that is:
$$\mathrm{IDF}(t, D) =  \log \frac{|D|}{|\{d \in D: t \in d\}|}$$ 

A further improvement to the syntactic characterization of terms was
introduced in the work on machine learning for Sledgehammer
\cite{KuhlweinBKU13}, where walks through term graphs were
considered. Adding such features again allows the abstraction of the global
term structure and better sharing of such automatically created concepts/features between the statements.

\subsection{ENIGMA Syntactic Features}
\label{enigmafeatures}

Syntactic features are also heavily used by ENIGMA systems, where clauses are
represented by finite numeric \emph{feature vectors}.
ENIGMA (\emph{\textbf{E}fficient
  Lear\textbf{n}ing-Based \textbf{I}nference \textbf{G}uiding
  \textbf{Ma}chine}) 
is a state-of-the-art machine learning guidance system for 
the ATP {\sc E}~\cite{SCV:CADE-2019}.
In the first ENIGMA~\cite{DBLP:conf/mkm/JakubuvU17}, the feature vector is
constructed by traversing the clause syntax tree and collecting all top-down
oriented symbol paths of length $3$.
For example, given the unit clause $\textit{plus}(X, nul) = X$, we obtain triples
$(\oplus,=,\textit{plus})$,
$(\oplus,=,X)$,
$(=,\textit{plus},X)$, and
$(=,\textit{plus},nul)$, where $\oplus$ signifies the root node of the syntax
tree of positive literals.
Additionally, to abstract from variable names and to deal with possible
collisions of Skolem symbols, all variables are replaced by a special
name $\odot$ and all Skolem symbols by $\ocoasterisk$.
After this renaming, the triples contain only the symbols from the problem
signature $\Sigma$ and $4$ special symbols
$\{\oplus,\ominus,\odot,\ocoasterisk\}$, where
$\ominus$ is used as the root node of negative literals.
This allows exhaustive enumeration of all possible triples, assigning each
triple a unique number smaller than $(|\Sigma|+4)^3$.
This number is used as an index in the feature vector, and the vector value
specifies the number of occurrences of the corresponding triple in the clause.

While the first version of ENIGMA yielded encouraging results, it was not yet
ready to scale to benchmarks with larger signatures.
This led to the second ENIGMA~\cite{JakubuvU18} with enhanced feature vectors.
Instead of an exhaustive enumeration of all possible symbol triples, only the
triples appearing in the training data were enumerated.
This significantly reduced the vector length as many triples do not appear in
the provided training data, and many can not appear at all, for example,
$(=,=,=)$.
This enumeration must be stored together with the trained model.
The second ENIGMA additionally introduced the following additional clause
features.

\begin{description}
\item[Count Features] extend the feature vector with the clause
   length, and the counts of positive/negative literals.
   Moreover, for each symbol $f$ we added the number of occurrences of $f$ in
   positive/negative literals, together with the maximal depth of $f$ in
   positive/negative literals.
   Count features allowed us to drop the clause length multiplier $\gamma$ and
   to use the model prediction directly as the clause weight.
\item[Horizontal Features] provide a more accurate representation of clauses by
  feature vectors.
  For every term $f(t_1,\dots,t_n)$, a new feature $f(s_1,\dots,s_n)$ was
  introduced, where $s_i$ is the top-level symbol of $t_i$.
  The number of occurrences of each horizontal feature is stored in the feature
  vector.
  Again, only the horizontal features that appear in the training data are
  considered.
  For example, the unit clause $\textit{plus}(X, nul) = X$,
  yields horizontal features
  $=(\textit{plus},\ocoasterisk)$ and
  $\textit{plus}(\ocoasterisk,\textit{nul})$,
  each occurring once in the clause.
\item[Conjecture Features] embed the conjecture to be proved in the feature
   vector.
   The first ENIGMA simply recommended the same clauses independently of the
   conjecture being proved.
   In the second ENIGMA, conjecture features were appended to the vector,
   making the vector twice the size.
   Thusly, the second ENIGMA was able to provide goal specific predictions,
   which was essential for experiments on Mizar problems which are much more
   heterogeneous than AIM benchmarks.
\end{description}

Another important step towards large data support in ENIGMA, was the implementation of \emph{feature
hashing}.
This significantly reduced the feature vector size.
ENIGMA uses a generic purpose string hashing function \emph{sdbm}.\footnote{
Given the (code of the) $i$-th character $s_i$:
$h_i = s_i + (h_{i-1} \ll 6) + (h_{i-1} \ll 16) - h_{i-1}$
with $h_0 = 0$.
}
Each feature is represented by a unique string identifier, for example,
$(=,\textit{plus},X)$ becomes ``\verb+|=|plus|*|+'', and
$\textit{plus}(\ocoasterisk,\textit{nul})$ becomes ``\verb+.plus.*.nul.+''.
This string is passed through the hashing function, computed with a
fixed-length data type representation ($64$ bit unsigned).
The string hash modulo the selected \emph{hash base} is used as the feature
index.
The hash base is intended to directly limit the vector size, at the price of
occasional feature collisions.

In order to abstract from specific symbol names also in the context of
features, the next ENIGMA~\cite{JakubuvCOP0U20} introduced a very simple method
of \emph{symbol anonymization}.
During the extraction of clause features, all symbol names are replaced by
symbol arities, keeping only the information whether the symbol
is a function or a predicate.
For example, the binary function symbol \textit{plus} becomes simply
\textit{f2}, the ternary predicate symbol \textit{ite} becomes
\textit{p3}, and so on.
In this way, a decision tree classifier does not depend on symbol names, for
the price of
symbol collisions.
While this rather trivial symbol anonymization was initially implemented mainly
as a baseline for the graph neural networks, it performed surprisingly
well in practice and it became a useful ENIGMA option.
The symbol collisions reduce the size of training data, which is a favorable
side effect of name anonymization.

\subsection{More Semantic Features}
\label{sf}

The syntactic structure of mathematical statements does not always easily capture %
their intended meaning. For examples, the syntactic features of two complex formulas $\phi$ and $\lnot\phi$ differ only very little, making them very close in various feature metrics. On the other hand, in the semantic Tarski-Lindenbaum algebra, these formulas are each other's complements, i.e., in some sense they are as distant as possible.
To accommodate this, %
mechanisms for extracting features that are based on the meaning of formulae
have been proposed.

The most common way of specifying the semantics of logical expressions in first-order logic is by considering which \emph{models satisfy the expressions}. However, a formula may have many satisfying models. Thus, only \emph{interesting} models %
that distinguish between formulas, are useful for characterization purposes. MaLARea-SG1~\cite{US+08} used finite model finders to continuously search for countermodels for problems with too few recommended premises. %
The notion of model's interestingness is thus defined dynamically, based on the current state of the learned knowledge (the trained premise selector), with which it co-evolves. 
This led to considerable improvement of MaLARea's performance.

Another approach to obtain more semantic features considers the
\emph{unifiability of two formulas}. For most first-order automated reasoning
calculi, the unifiability of formulas correlates with their use as
principal formulas of an inference step. %
This means that first-order indexing structures~\cite{KaliszykUV15} can be used as another
source of more semantic features. The nodes of indexing structures, such as
a substitution tree or discrimination tree, correspond to sets of
unifiable formulas present in the given reasoning problem, while paths
correspond to similar term structures. Thus, features corresponding to
these nodes %
can characterize whether the given
formulas can be combined %
for inferencing within a given calculus.

\emph{Latent semantic analysis} (LSA) \cite{DeerwesterDLFH90} is a method that creates
low-rank vectors (\emph{embeddings}) characterizing terms and
documents based on the terms' co-occurrence in the documents. LSA-based
features were used with some success in premise selection systems \cite{DBLP:conf/cade/KaliszykUV14} and
hammers \cite{KaliszykU15a}. LSA-based embeddings precede related more recent methods,
which use the same idea of characterizing words by their context, such
as Word2Vec \cite{mikolov2013efficient} and neural embeddings.

\subsection{Characterization Using Neural Networks}
\label{sec:CharNeu}

Already the early tree neural networks by Goller and Kuchler attempt
to learn the representation of the symbols as neural sub-networks.
More recently, starting with Word2Vec (a shallow neural network), a number of neural approaches have been experimented with for obtaining useful sentence and term embeddings. %

For example, encoder-decoder neural frameworks attempt to summarize an input text (e.g., in English) in a vector that can be then decoded into suitable output (e.g., French text with corresponding meaning).
A recent encoder-decoder approach was developed by Sutton et al. \cite{AllamanisCKS17}.
It focused on developing an embedding capturing
semantic equivalence, i.e., formulas that are negations of each other
are adequately distinguished. This research direction is connected to
the work outlined in Section~\ref{sec:neursynth}, where approximate
reasoning and semantically rich encoding are used to build systems for
various synthesis problems. %
Purga\l{} \cite{PurgalPK21} later developed an autoencoder for predicate logic formulas by training a network to decode a given formula's top symbol and its children.
Another example %
is~\cite{ZhangBKU23}, which builds a representation of tactics by their meanings, i.e., how the tactic transforms a given state into several subsequent states.  

An alternative to vectorization of symbolic expression as simple
sequences of symbols is to use neural architectures that capture
the structure of the expressions.
This includes the pioneering work on Tree NNs by
\cite{DBLP:conf/icnn/GollerK96}, capturing the recursive nature and tree-like structure of logical expressions.
Related investigation is done in~\cite{EvansSAKG18} for recognizing propositional logical
entailment through evaluation of formulas in `possible worlds', which is also related to the earlier model-based features in MaLARea.  A
related investigation~\cite{Chvalovsky19} models expressions using a
similar network architecture but instead aims to recognize properties
such as \textit{validity} through a top-down evaluation of the
expression. The model thus tries to approximate decompositional
reasoning.

An illustrative example of Tree NNs modeling symbolic expressions for
the purpose of guiding an automated reasoner is
\textit{ENIGMA-NG}~\cite{ChvalovskyJ0U19}. The authors provide a
vector embedding that associates each predicate (function) symbol with
a learned function $R^n\times \cdots \times R^n\rightarrow R^n$ where
the number of arguments matches the symbol's arity. Thus, the term
structure is encoded by the composition of these functions. Certain
expressions whose semantic content cannot be directly extracted from
the structure of the expression are grouped together. For example, all
variables share the same encoding function. Clauses, rather than being
considered as a composition of several instances of a binary \emph{or} function,
are handled by a separate RNN model capturing their set-like nature.

While Tree NNs capture the structure of symbolic expressions,
a few issues remain. For example, the encoding used by
\textit{ENIGMA-NG}~\cite{ChvalovskyJ0U19} requires the construction of
a learnable function for each \textit{symbol}, while all variables use a single learnable
function.
Graph neural networks (GNNs)~\cite{4700287} can provide an architecture able to abstract
away the names used for the representation of symbolic
expressions and, to some extent, distinguish variables
and their occurrences. Early uses of GNNs for premise selection, such
as FORMULANET~\cite{10.5555/3294996.3295038}, did not improve these
deficiencies but provided direction towards a more semantic-preserving
architecture. Olsak~\cite{OlsakKU20} then proposed a GNN-based name-invariant embedding 
using a hypergraph of clauses and terms, where none of the names appear and a symbol's meaning can only be inferred from the subgraph of its properties. While some loss of structure may
still occur, for example, $f(t_1,t_2,t_1)$ and $f(t_2,t_1,t_2)$ are
identically encoded, the embedding is a vast improvement over previous
approaches, naturally treating also clauses as sets.

Such embeddings allow the neural architecture to draw analogies
between different mathematical domains. Many algebraic operators are
associative, commutative, and/or distributive, yet they have different names.
The framework presented in~\cite{OlsakKU20}
would recognize the local graph structure as analogous regardless of the different names used.
While such neural
architectures provide better generalization and improved cross-domain predictions, compared to simpler methods such as decision trees their predictions may be too confident.
This problem was observed
when integrating the above-mentioned GNN architecture into the
inference selection mechanism of
\textit{plcop}~\cite{ZomboriUO21}. The authors introduced
\textit{entropy regularization} to normalize the model's confidence
and improve accuracy.

\section{Premise Selection}
The success of modern ATPs is partially due to
their ability to select a small number of facts that are
relevant to the conjecture. %
The act of
selecting these relevant facts is referred to as \textit{premise
  selection} \cite{AlamaHKTU14}, which can also be seen as a step in a more general abstraction-refinement framework \cite{HernandezK18}.
Without good premise selection,
ATPs  can be easily overwhelmed by a number of possible
deductions, which holds even for relatively simple conjectures. Non-learning ATP approaches include heuristics such as MePo and SInE, mentioned in Section \ref{hist}. 
In this section, we cover the typically more precise %
learning approaches. %

\subsection{k-Nearest Neighbors (k-NN)} 
This approach to selection requires a measure of distance, more appropriately referred to as a \textit{similarity relation}, between two facts. This similarity relation, in simple cases, is defined by the distance between two facts computed from a set of binary features. To increase precision, weights and a scaling factor are added to the features. Thus, the similarity relation between two facts is defined as follows:
\[s(a, b) = {\sum\nolimits_{\,f \in F(a) \cap F(b)}{w(f)}^{{\tau_1}}}\]
where $F$ is the feature vector, $w$ is the weight vector and $\tau_1$ is the scaling factor. 

\newcommand\factf{a}
\newcommand\factg{b}
\newcommand\facth{h}
 We then realize that if a proof relies on more dependencies,
  each of them is less valuable, so we divide by the number of dependencies. We additionally add a factor for the dependencies
  themselves. Given $N$ the set of the $k$ nearest neighbours, the relevance of fact $\factf$ for goal $g$ is:
\[
\left({\tau_2} \sum\nolimits_{\factg \in N \mid \factf \in D(\factg)\,}
\frac{s(\factg, g)}{\left|D(\factg)\right|}\right)
{
  \mathrel+
{\begin{cases}
  s(\factf, g) & \text{if } \factf \in N \\[-1\jot]
  0 & \text{otherwise}
\end{cases}}
}
\]

This approach %
was later extended to adaptive $k$, i.e., considering more or less neighbours depending on the requested number of best %
premises~\cite{KaliszykU15a}. %

\subsection{Naive Bayes}
\label{nb}

  Given a fact $a$ and a goal to prove $g$, we try to estimate the probability it is useful based
  on the features $f_1, ..., f_n$ of the goal $g$:
\begin{eqnarray*}
& P(\text{$a$ is relevant for proving $g$}) \\
= & P(\text{$a$ is relevant} \mid \text{$g$'s features}) \\
= & P(\text{$a$ is relevant} \mid f_1,\hdots,f_n) \\
\varpropto & P(\text{$a$ is relevant}) \Pi_{i=1}^n P(f_i \mid \text{$a$ is relevant}) \\
\varpropto & \# \text{$a$ is a proof dependency} \cdot \Pi_{i=1}^n \frac{\# f_i \; \text{appears when $a$ is a proof dependency}}{\# \text{$a$ is a proof dependency}} \\
\end{eqnarray*}
This approach can be %
improved by considering features not present in the goal when $a$ was used. %
An issue with considering the negative case is that there are too many possible features to take into account.
So, instead of looking at all features not present when $a$ was used, %
we only consider the so called \emph{extended features} of a fact, namely the features that do not appear and are related to those that do appear \cite{KuhlweinBKU13}. This allows considering the probabilities of features not appearing in the goal while a dependency was used. 
These probabilities can be computed/estimated efficiently. Early versions of MaLARea used the %
implementation in the SNoW~\cite{SNoW} toolkit. %
Kaliszyk later \cite{KaliszykU13a,KaliszykU14} implemented a custom naive Bayes %
for Flyspeck and other experiments. %

\subsection{Decision Trees}

A \emph{decision tree} is a binary tree with
nodes labeled by conditions on the values of the feature vectors. Such trees and their \emph{ensembles} are today among the strongest ML methods, which can also be very efficient.
Initially, F\"{a}rber~\cite{FarberK15} improved on k-NN
approaches~\cite{KaliszykU13a} with \emph{random forests} (ensembles of decision trees) that used k-NN as a secondary classifier.
With further modifications,
random
forests can also be applied directly as,
e.g., in~\cite{PiotrowskiMA23} where they are used for premise
selection in Lean.  %
\emph{Gradient-boosted trees}, as implemented by
XGBoost~\cite{DBLP:conf/kdd/ChenG16} and
LightGBM~\cite{DBLP:conf/nips/KeMFWCMYL17} %
are useful both for premise selection
and for efficient ATP guidance as
discussed in the next section. Since they work in a binary (positive/negative) setting, they require (pseudo-)negative examples for training.
Piotrowski's \emph{ATPBoost}~\cite{PiotrowskiU18} defined an infinite MaLARea-style loop that interleaves their training with proving, producing increasingly better positive and negative data for the training.
Quite surprisingly, \emph{hashing} the large
number (over millions) of sparse symbolic features (such as term
walks) into much smaller space (e.g. 32000) allowed efficient training
of these toolkits over very large libraries such as full MML, with
practically no performance penalties~\cite{ChvalovskyJ0U19,jakubuv_et_al:LIPIcs:2019:11089}.

\subsection{Neural Methods}
\label{nnm}
The first analysis of the performance of deep neural networks for premise
selection was done by the \textit{DeepMath} \cite{IrvingSAECU16} project. Despite requiring significantly more resources, there was only limited improvement over the simpler methods. However, the work also employed the above mentioned \emph{binary setting}, where conjectures
and the potential premises were evaluated together. %
This allows to meaningfully evaluate
premises that have not been seen yet. %
Methods such as k-NN and naive Bayes typically do not allow that without further data-augmentation tricks.\footnote{Such as adding for each premise as a training example its provability by itself, which has indeed been used from the beginnings of ML-based premise selection~\cite{Urb04-MPTP0,Urban06}. Data augmentation in general is another very interesting AI/TP topic, see e.g. \cite{DBLP:journals/jsc/KaliszykU15}.}
As logical formulas naturally have a tree structure, graph neural networks were soon proposed for premise
selection \cite{10.5555/3294996.3295038} %
(see also Section~\ref{sec:CharNeu}).
Other alternatives such as graph sequence models \cite{abs-2303-15642}, directed graph networks \cite{RawsonR20},
recurrent neural networks \cite{PiotrowskiU202} and transformers \cite{UrbanJ20} have been experimented with, allowing to take into account also dependencies among the premises \cite{PiotrowskiU202}.

The signature independent GNN described in~\cite{OlsakKU20} seems to be the strongest method today, based on a recent large evaluation of many methods over the Mizar dataset \cite{JakubuvCGKOP00U23}. The study also demonstrates the role of \emph{ensembles and portfolios} of different premise selection methods. An interesting phenomenon observed in \cite{UrbanJ20} is that some of the neural methods provide a smooth transition between premise selection and \emph{conjecturing}, i.e., proposing so far unproved ``premises'' (intermediate conjectures) that split the problem into two \cite{GauthierKU16}. Such neural conjecturing methods are further explored e.g. in \cite{PiotrowskiU20,Rabe0BS21,abs-2210-03590,GauthierOU23,JohanssonS23}.
\section{Guidance of Saturation-based ATPs}
Several decisions that ATPs need to regularly perform are very similar to premise selection. 
The selection of the next clause in saturation-based provers or %
of the next extension step in a connection tableaux calculus both %
correspond to the selection of relevant %
premises for a given conjecture. For this reason, many ATP guidance techniques are motivated by premise selection, with the additional caveat that efficiency needs to be much more taken into account.

Arguably the most important heuristic choice point in saturation-based
theorem proving is \emph{clause selection}. It is the procedure for
deciding, at each iteration of the main proving loop, which will be
the next clause to activate and thus participate in generating
inferences. A perfect clause selection---obviously impossible to
attain in practice---would mean selecting just the clauses of the
yet-to-be-discovered proof and would thus completely eliminate search
from the proving process. Although state-of-the-art clause selection
heuristics are far from this goal, experiments show that even small
improvements in its quality can have a huge impact on prover
performance \cite{DBLP:conf/cade/SchulzM16}. This makes clause
selection the natural main target for machine-learned prover guidance.

The central idea behind improving clause selection by ML, which goes
back at least to the early work of Schulz
\cite{Sch95,DS1996b,Schulz:KI-2001}, %
is to learn from successes. One trains a binary classifier for
recognizing as positive those clauses that appeared in previously
discovered proofs and as negative the remaining selected ones. In
subsequent runs, clauses classified positively are prioritized for
selection.

Particular systems mainly differ in 1) which base prover they attempt
to enhance, 2) how they represent clauses for the just described
supervised learning task, and 3) which ML technique they use. When
aiming to improve a prover in the most realistic, i.e., time
constrained, setting, there are intricate trade-offs to be made
between faithfulness of the chosen representation, capacity of the
trained model and the speed in which the advice can be learned and
retrieved. State-of-the-art provers are tightly optimized programs and
a higher-quality advice may not lead to the best results if it takes too
long to obtain.\footnote{This is not only because search and/or backtracking is
an indispensable part of any reasoning which is not purely memorized, but
also because self-learning systems that use faster guidance will produce more data to learn from. In today's ML, a slightly weaker learner trained on much more data will often be better than a slightly stronger learner trained on much less data. If a self-learning system uses a slightly weaker but much faster learner/predictor to do many iterations of proving and learning, the self-learning system will in a fixed amount of time produce much more data and its ultimate performance will thus be higher.}

For example, the ENIGMA (\emph{\textbf{E}fficient
  Lear\textbf{n}ing-Based \textbf{I}nference \textbf{G}uiding
  \textbf{Ma}chine}) system, extending the ATP {\sc
  E}~\cite{SCV:CADE-2019}, started out with easy-to-compute syntactic
clause features (e.g., term-walks, see Section~\ref{enigmafeatures}) and a simple but very
\emph{efficient} linear classifier
\cite{DBLP:conf/mkm/JakubuvU17,JakubuvU18}. At the same time, Loos et
al. \cite{LoosISK17}, %
first experimenting with integrating state-of-the-art neural networks
with E, discovered their models to be too slow to simply replace the
traditional clause selection mechanism. A similar phenomenon appears in
ML-based guidance of tactical ITPs (Section \ref{tactic}), where the relatively simple but
fast predictors used in systems like TacticToe \cite{GauthierKUKN21}
(HOL4) and Tactician \cite{BlaauwbroekUG20} (Coq) are hard to beat by
more complicated but significantly slower neural guidance in a fair
comparison \cite{graph2tac}.
In later versions of ENIGMA, these trade-offs were further explored by experimenting with
gradient boosted trees and tree neural networks over term structure
\cite{ChvalovskyJ0U19}, and finally with graph neural networks
\cite{JakubuvCOP0U20}. In the meantime, Deepire \cite{000121a,Suda21},
an extension of the prover Vampire \cite{Vampire} used tree neural
networks but over clause derivation structure. 
In the rest of this
section, we compare these and other systems also from other angles as
we highlight some notable aspects of this interesting technology.

\paragraph{Integrating the learned advice.} There are several possible ways in which the trained model $M$ can be used to improve clause selection in a guided prover. Typically, one seeks to first turn the advice into a total order on clauses, e.g., for maintaining a queue data structure for a quick retrieval, and, in a second step, to somehow combine this order with the original clause selection heuristic.

In the most general case of a black-box binary classifier, $M$ only suggests whether a clause is good (1) or bad (0). The first ENIGMA \cite{DBLP:conf/mkm/JakubuvU17} used a linear combination of the $M$'s classification and clause's length to come up with an overall score for sorting clauses. Not fully relying on $M$ makes the advice more robust, in particular, it mitigates the damage from having large erroneously positively classified clauses. 
Another sensible strategy, adopted, for example, by ENIGMA-NG \cite{ChvalovskyJ0U19}, is to have all positively classified clauses precede the negatively classified ones and to use clause generation time-stamps (sometimes referred to as \emph{clause age}) as a tiebreaker on clauses in each of the two classes. This amounts to saying ``prefer clauses suggested by $M$ and among those apply the FIFO rule: old clauses first before younger ones''.
Finally, even a binary classifier often internally uses a continuous domain of assigned values (often referred to as \emph{logits}) which are only turned into a binary decision by, e.g., getting compared against a fixed threshold. This is for instance true for models based on neural networks. ENIGMA anonymous \cite{JakubuvCOP0U20}, as well as the system by Loos et al.~\cite{LoosISK17}, used a simple comparison of clause logits for ordering their neural queue.

A clause queue ordered with the help of $M$ (in one of the ways just
described) can of course be used to simply replace any original clause
selection heuristic in the prover. However, it is often %
better to
combine the original heuristic and the learned one
\cite{ChvalovskyJ0U19,000121a}. The simplest way of doing this is to
alternate, in some pre-selected ratio, between selections from the
various clause queues (one of them being the one based on $M$).
A \emph{layered clause selection} mechanism, which lead to the best performance of Deepire \cite{000121a},
applies the full original heuristic to clauses classified positively by $M$ and alternates this with 
the original heuristic applied to all clauses. An advantage of this approach is that it allows for 
a \emph{lazy evaluation} trick \cite{000121a}, under which not every clause available for selection
has to immediately get evaluated by the relative expensive model $M$.
Very recently, in the context of the iProver (instantiation-based)
system, guidance using only the trained GNN model $M$ managed to
outperform the combination of $M$ with the original heuristics
\cite{ChvalovskyKPU23}. One of the reasons for that is an improved
learning from \emph{dynamic} proof data, taking into account more
proof states than the traditional ENIGMA-style training that only
learns from the final proof states.

\paragraph{Signature (in-)dependence.} First-order clauses, as
syntactic objects, are built from predicate and function symbols
specified by the input problem's \emph{signature}. A representation of
clauses for machine learning may consider this signature fixed, which
is often natural and efficient, but ultimately ties the learned
guidance to that signature. E.g., if we train our guidance on problems
coming from set theory, the model will not be applicable to problems
from other domains.

The early pattern-based guidance by Schulz for E was already signature
independent, as well as, e.g., the early concept-alignment methods
designed for transfer of knowledge between ITP libraries by Gauthier
and Kaliszyk~\cite{GauthierK19}.  The ENIGMA systems
\cite{DBLP:conf/mkm/JakubuvU17,JakubuvU18,ChvalovskyJ0U19} relied
initially on the fixed signature approach. ENIGMA Anonymous
\cite{JakubuvCOP0U20} started %
to abstract from
specific symbol identities and replace them with arity abstractions of
the term walks and property invariant neural embeddings
\cite{OlsakKU20}, opening doors to knowledge
transfer. %
A detailed comparison of these ENIGMA abstraction mechanisms with the
earlier ones by Schulz, Gauthier and Kaliszyk is discussed
in~\cite{JakubuvCOP0U20} (Appendix B).

An interesting approach from this perspective, is taken by Deepire
\cite{000121a,Suda21}, which uses recursive neural networks for
classifying the generated clauses based solely on their derivation
history.  Thus Deepire does not attempt to read ``what a clause
says'', but only bases its decisions on ``where a clause is coming
from''. This makes it trivially independent on problem signature,
however, it still relies on a fixed initial axiom set over which all
problems are formulated.

\paragraph{Building in context.} A clause $C$ useful for proving conjecture $G_1$ can be completely useless for proving a different conjecture $G_2$. While a great prover performance boost via learning can already be achieved \emph{without} taking the conjecture context into account \cite{Suda21} (and, indeed, the standard clause selection heuristics are of this kind), many systems supply some representation of the conjecture as a secondary input to their model to improve the guidance \cite{LoosISK17,JakubuvU18,ChvalovskyJ0U19}.

Another, more subtle, kind of context is the information of how far the prover is in completing a particular proof. Intuitively, selecting a certain clause could only make sense if some previous clauses have already been selected. This is explored by ENIGMAWatch \cite{GoertzelJU19}, where a proof state is approximated by a vector of \emph{completion ratios} of a fixed set of previously discovered proofs. A system called TRAIL goes further and allows every processed clause to influence the score of any unprocessed clause through a multiplicative attention mechanism \cite{CrouseAMWCKSTWF21}. This level of generality, however, is computationally quite costly, and TRAIL does not manage to improve over plain E prover \cite{SCV:CADE-2019} under practical time constraints.

\paragraph{Looping and reinforcement.} Once we train a model that
successfully improves the performance of the base prover, we can
collect even more proofs to train on and further improve our guidance
in a subsequent iteration. Such re-learning from new proofs, as
introduced by MaLARea \cite{US+08}, constitutes a powerful technique
for tapping the full potential of a particular guidance architecture
\cite{jakubuv_et_al:LIPIcs:2019:11089,Suda21}.

Iterative improvement is also at the core of the reinforcement
learning (RL) approach \cite{DBLP:books/lib/SuttonB98}, which builds
on a different conceptual framework than the one we used so far (e.g.,
agent, action, state, reward -- see also Section \ref{tableau}), but in the context of saturation-based
provers gave so far rise to systems of comparable design
\cite{CrouseAMWCKSTWF21,DBLP:conf/icml/AygunAOGMFZPM22}. %
The main reason for this is that even systems based on RL reward (and
reinforce the selection of) clauses that appeared in the discovered
proofs. The alternative of only rewarding the final proof-finishing
step and letting the agent to distinguish the good from the bad
through trial-and-error would be prohibitively expensive.

\paragraph{Beyond clause selection.} Although clause selection has
been the main focus of research on this front, there are many other
ways in which machine learning can be used to improve saturation-based
provers. For example, it is possible to predict good term orderings
for the underlying superposition calculus \cite{Bartek021} or symbol
weights \cite{Bartek023}.  In the later case we still aim to influence
clause selection via clause weight (one of the standard heuristics),
but only indirectly, in an initialization phase, which comes before
the proof search starts.
Another class of approaches, which we only briefly mention here, use
ML-style techniques for %
synthesis of ATP strategies and suggesting good targeted strategies or
strategy schedules based on input problem features
\cite{blistr,SchaferS15,KuhlweinU15,JakubuvU18a,DBLP:conf/mkm/HoldenK21}.

ML-based approaches have also been used to prevent interactions
between unsuitable clauses \cite{GoertzelCJOU21}, and in
general to design iterative algorithms (\emph{Split \& Merge}) that
repeatedly split the problems into separate reasoning components whose
results are again merged after some time \cite{ChvalovskyJOU21}. Such
algorithms can also be seen as soft/learnable alternatives to ATP
methods such as \emph{splitting} \cite{DBLP:conf/cav/Voronkov14}, and to manual design of
theory procedures and their combinations in SMT.

\section{Guidance of Tableaux and Instantiation-based ATPs}
\label{tableau}
While the saturation-based ATPs are today the strongest, 
the proofs produced by %
such systems are often at
odds with human intuition. Tableaux-based provers produce proofs
closer in argumentation %
to human reasoning; this is likely due
to the case-based style of the typical tableaux proof system.
Essentially, such systems are performing a sort of model elimination
rather than a search for contradiction. An advantage of such
approaches is that the proof state is compactly represented, and it is easier to control 
the number of possible
actions in comparison to saturation provers, where the number of
mutually resolvable clauses can grow quickly.  This has resulted in a lot of recent research in adding ML-based guidance in
particular to the \textit{connection tableaux
  calculus}~\cite{DBLP:books/sp/17/OttenB17}.

First, the MaLeCoP (Machine Learning Connection Prover)
system~\cite{UrbanVS11} used an external and relatively slow
evaluation method to select the extension steps in the leanCoP ATP
\cite{OB03}. This showed that with good guidance, one can avoid 90\%
of the inferences. This was made much faster by integrating an
efficient sparse naive Bayes classifier in an ML-guided OCAML
re-implementation of leanCoP (FEMaLeCoP)~\cite{KaliszykU15}. As an alternative to the direct selection
of \textit{extension steps}, Monte Carlo simulations can be used to
select the promising branches (MonteCoP) \cite{FarberKU21}.  A major progress
was obtained by removing the \emph{iterative deepening} used by default in leanCoP, and instead using an Alpha-Zero-like architecture for
guiding connection tableaux in \emph{rlCoP}~\cite{KaliszykUMO18}. Reinforcement learning
of \emph{policy} (action, i.e. inference, selection) and \emph{value} (state, i.e., partial  tableaux, evaluation) and their use for intelligent subtree
exploration yielded after several iterations of proving and learning on a training set a system that solves 40\% more test problems
than the default leanCoP strategy .  The method
still produces rather short proofs and policy guidance methods were
later investigated for proofs with thousands of inference steps
\cite{ZomboriCMKU21}. Extensions of connection tableaux, such as lazy
paramodulation, can also be directly guided using similar methods
\cite{RawsonR21}. Further improvements were recently achieved by
integrating signature-independent graph neural networks (Section
\ref{sec:CharNeu}), e.g. \cite{OlsakKU20,ZomboriUO21}.

Instantiation-based and SMT (Satisfiability Modulo Theories)
systems such as iProver~\cite{Korovin08}, CVC (CVC4/cvc5)~\cite{cvc5} and Z3~\cite{z3} combine the use of SAT
solvers for checking ground (un)satisfiability with various methods
for producing suitable ground instantiations of the first-order
problems. This approach goes back to the early days of ATP
(Section~\ref{hist}) and methods such as Gilmore's~\cite{gilmore1960proof}, however they have recently become much
more relevant thanks to today's powerful CDCL~\cite{DBLP:conf/iccad/SilvaS96} based SAT solving and other
calculus improvements.

In SMT, ML has so far been mainly used for tasks such as portfolio and
strategy
optimization~\cite{10.1007/978-3-030-72013-1_16,10.1007/978-3-030-80223-3_31,BalunovicBV18}.
More recent
work~\cite{janota-sat22,el2021methodes,blanchette:hal-02381430} has
also explored fast non-neural ML guiding methods based on decision
trees and manual features. Very recently, the first neural methods for
guiding the cvc5 SMT system have started to be
developed~\cite{cvc5gnn24}. Due to the large number of possible instantiations
in such settings, this is typically more involved than guiding the
clause selection as in the saturation based systems. Interestingly,
iProver's instantiation-based calculus is also using the given clause
loop, and an efficient neural guidance of its Inst-Gen~\cite{DBLP:conf/birthday/Korovin13} procedure has
recently led to doubling of its performance on the Mizar
corpus~\cite{ChvalovskyKPU23}.

\section{Tactic Based ITP Guidance}
\label{tactic}

In most interactive theorem provers, proofs are written using meta-programs consisting of \textit{tactics}. A tactic can analyze the current state of the proof and generate a sequence of kernel inference steps or a partial proof term to advance the proof. Actions performed by tactics can range from simple inference steps, to decision procedures, domain-specific heuristics, and even a generic proof search. 
As an alternative to ATP guidance of low-level steps, it is possible to recommend such tactics and explore the proof
space using them. Recommendations are made by analyzing existing tactical proofs (either written by users or found automatically) and learning to predict which tactic performs a useful action on a given proof state. This \emph{mid-level} guidance task falls anywhere between premise selection and the ATP guidance, allowing ML methods that may be slower than those used for ATPs.

\subsection{Overview}
Early systems in this field include ML4PG \cite{abs-1212-3618} for Coq, which gives tactic suggestions by clustering together various statistics extracted from interactive proofs, without trying to finish the proofs automatically.
SEPIA \cite{GransdenWR15} provides tactic predictions and also proof search for Coq, which is however only based on tactic traces without considering the proof state. 
The first 
system that considers proof states is TacticToe \cite{GauthierKU17}, which uses k-NN selection to predict the most
likely tactics that complete a goal in the HOL4 proof assistant. Its later versions combined the prediction
of promising tactics with Monte-Carlo tree search giving a very powerful method (66\% of the library proved) for proof automation \cite{GauthierKUKN21}.

Similar systems have been created for Coq (Tactician \cite{BlaauwbroekUG20}) and HOL Light (HOList \cite{BansalLRSW19}),
as well as frameworks for the exploration of the tactical proof space in Lean  \cite{abs-2306-15626}.  
The early PaMpeR~\cite{NagashimaH18} system for Isabelle also considers the proof states, however, it only recommends one command for each proof state, leaving its execution to the user (i.e., it lacks the search component). 
Further (sometimes experimental) systems for Coq include GamePad \cite{huang2018gamepad} and CoqGym \cite{pmlr-v97-yang19a}, using deep neural networks and slow (600s) evaluation mode, as well as Proverbot9001 \cite{abs-1907-07794}, which was shown to perform well on CompCert. Further Coq-oriented systems (TacTok, Diva, Passport) \cite{tactok,diva,passport} are based on the
dataset provided by CoqGym.
For the Lean proof assistant, further systems include
LeanDojo \cite{yang2023leandojo} and \cite{DBLP:conf/nips/LampleLLRHLEM22}.

\subsection{Advantages of Tactic-based ITP Guidance}
There are several advantages associated with using tactics for proof search.
    \paragraph{Adaptivity.} Tactics provide a higher level and more flexible base for proof search. Generating kernel inference steps or partial proof terms directly is a task that needs high precision. On the other hand, tactics are often adaptive and can perform a sensible high-level action, such as search or decision procedures, on a wider variety of proof states. 
    \paragraph{Specialization.} In highly specialized branches of mathematics, tactics are often specialized to the domain by experts. That is, when the default set of tactics provided by a proof assistant is not satisfactory, a new set of tactics may be written by end-users. This is particularly useful when the mathematical domain requires a deep embedding of a custom logic.
    An example of this is the Iris separation logic framework~\cite{DBLP:journals/jfp/JungKJBBD18}  for Coq. It includes a set of tactics specifically crafted for working with separation logic. Another example is the CakeML project~\cite{cakeml14} for HOL using custom tactics, which provides a formally verified compiler for the ML language. 
        Some tactic-based proof search methods, such as Tactician's $k$-NN model for Coq~\cite{BlaauwbroekUG20} can learn to use new tactics in real time, which has proven rather powerful~\cite{graph2tac}.

\subsection{Challenges in Tactic-based ITP Guidance}

\paragraph{Potential Incompleteness.} Contrary to performing basic inference steps for a logic (as e.g. in the ATP calculi), tactics might not
represent a minimal and complete set of inference rules.
A set of tactics may not be guaranteed to be able to prove every theorem.\footnote{This is a theoretical concern that usually is not a problem in practice.}

\paragraph{Overlap in functionality.}
The actions taken by different tactics may have a high degree of overlap.
Systems like Proverbot9001~\cite{abs-1907-07794} and Tactician~\cite{BlaauwbroekUG20} attempt to mitigate tactic overlap by \textit{decomposing} and \textit{normalizing} tactic expressions, attempting to eliminate duplicate tactics.

In TacticToe, the \emph{orthogonalization} process is introduced to eliminate redundant tactics~\cite{GauthierKUKN21}. In more detail,
TacticToe maintains a database of goal-tactic pairs used for training and this database is subject to the orthogonalization process which is
intertwined with the learning.
Orthogonalization works as follows: First,
each time a new tactic-goal pair (t, g) is
extracted from a tactic proof and about to be recorded in the database, we consider if there already 
exists a better tactic for the goal g in the database. To this end, we organize a competition between the k
tactic-goal pairs that are closest\footnote{He we already use the learned notion of proximity on the database constructed so far.}
to the pair (t,g) (including it). The winner (which is ultimately stored in the database and trained on)
is the tactic that subsumes the original tactic t
on the goal g and that appears in the largest number of tactic-goal pairs in the database.
As a result, already successful tactics with a large coverage
are preferred, and new tactics are considered only if they provide a different contribution.

    \paragraph{Diverse Tactic Behavior.} Due to the diverse range of tactic behaviors it can be difficult to tune proof search to appropriately exploit each class of tactics. For example, while simple tactics are executed rather quickly, more sophisticated tactics may require multiple seconds of execution time before their action completes. Deciding which tactic to execute and the appropriate allocation of resources to it is a major challenge.
    \paragraph{Tactics Are Designed for Humans.} Because tactics are designed for use by humans, they usually come with a complex language of tactic combinators, higher-order language features, and syntactic sugar.
Typically, in machine learning and reinforcement learning we however prefer to have a relatively simple set of actions (commands in the ITP setting). Therefore one would like to decompose the proof scripts into sequences (or more generally trees) of simpler commands. This is a difficult preprocessing step.

\paragraph{Representation.} Simpler tactic-based systems make predictions solely based on surface-level syntax features of proof states. This makes it difficult to extract deeper knowledge about the previously introduced concepts. Also, straightforward neural encodings are typically insufficient when new concepts and lemmas are added on the fly (which is very common in ITP), because it is expensive to always adapt a large neural model after the addition of such new items.
The most recent work on Graph2Tac \cite{graph2tac} learns representations of all definitions in its dataset and can generate representations for unseen definitions on-the-fly. This provides more accurate representations of proof states and tactic arguments, resulting in a 50\% improvement over baselines that do not incorporate such background information.

\section{Related Symbolic Classification Problems}
\label{sec:neursynth}
The majority of investigations discussed above concern the
classification of symbolic expressions with the goal of informing a
symbolic system, which of a variety of actions is most likely to
result in success. A few exceptions were discussed in
Section~\ref{sec:CharNeu}, for
example~\cite{EvansSAKG18,Chvalovsky19}. Both papers present an
embedding and a neural architecture solving a classification problem
but with no intention of integration within a symbolic system. While
this positions such investigations quite far from the core topic of
this Survey, note that these works motivated the approach presented
in~\cite{ChvalovskyJ0U19} where the authors used a tree NN to guide
clause selection. Thus, it is likely that future developments
improving precise selection and guidance will be motivated by
approaches developed for other symbolic classification problems.

Relatively recent investigations have considered the introduction of
probabilities into logic
programs~\cite{DBLP:journals/tplp/FierensBRSGTJR15}, i.e., some facts
are associated with probability. Such logic programs allow the
introduction of predicates whose definition is a neural network. The
authors of \textit{DeepProbLog}~\cite{ManhaeveDKDR18,ManhaeveDKDR21},
introduce an approach to train so-called \textit{neural predicates}
within the context of a probabilistic logic program. In some sense,
this can be viewed as a form of \textit{symbolic guidance} of the
training procedure of a statistical model. While the majority of this
Survey, including the section on neural classification, focuses on
guidance, a few investigations have considered \textit{end-to-end}
neural theorem proving~\cite{Rocktaschel017}, that perform a soft
unification operation over a vector representation of the proof state
and atoms to unify. Investigations have also considered the use of
\textit{generative adversarial networks} to train a prover and a
teacher simultaneously with the goal of teaching the prover to solve
the problems presented by the
teacher~\cite{DBLP:conf/flairs/PurgalK22}.

ILP is a form of symbolic machine learning whose goal is to derive
explanatory hypotheses from sets of examples (denoted $E^+$ and $E^-$)
together with background knowledge (denoted
$\mathit{BK}$)~\cite{DBLP:journals/jair/CropperD22}. Essentially, it
is a form of inductive synthesis. Early approaches to providing a
statistical characterization of ILP include
\textit{nFOIL}~\cite{DBLP:conf/aaai/LandwehrKR05}, which models the
search procedure and stopping conditions of
FOIL~\cite{DBLP:journals/ml/Quinlan90} using Naive Bayes. Essentially,
nFOIL attempts to maximize the probability that a hypothesis covers
the examples. The stopping condition is the score function reaching a
user-defined threshold. A similar approach was investigated by
integrating Kernel methods and FOIL
\cite{DBLP:conf/aaai/LandwehrPRF06}.

More recently, Evans et al.~\cite{DBLP:journals/jair/EvansG18}
introduced \textit{$\delta$ILP} which considers ILP as a
satisfiability problem where each proposition denotes a pair of
clauses constructable using the \textit{background knowledge}. These
clause pairs are used as the definition \textit{predicated
  template}. The hypothesis consists of a user-defined number of
predicated templates. Essentially, a propositional model of this SAT
problem is a hypothesis. As a next step, the authors introduce a type
of soft inferencing used to compute the evaluations of the
propositions composing the SAT problem. Investigations based on the
$\delta$ILP have considered a hierarchical structure of
templating~\cite{pmlr-v162-glanois22a} and massive predicate
invention~\cite{DBLP:journals/corr/abs-2208-06652}. Recent work by
some of the Authors of $\delta$ILP illustrates that such
\textit{neuro-symbolic} system can be interleaved with a symbolic
solver, allowing them to provide feedback during
training~\cite{EVANS2021103521}.

While $\delta$ILP requires learning a propositional model for a SAT
encoding of an ILP instance through soft inferencing, the framework is
not directly viable as a general \textit{neural} SAT
solver. NeuroSAT~\cite{SelsamLBLMD19} embeds a SAT instance into a GNN
that learns which propositions are required for satisfiability. A more
recent approach to neural SAT solving is the
SATNet~\cite{WangDWK19,NEURIPS20200ff8033c,NEURIPS2021ad7ed5d4} where
satisfiability is formulated as a semidefinite program.

\section{Conclusion}

In this work, we discussed the main contemporary areas of combining
automated reasoning and especially automated theorem proving with
machine learning. This includes the early history, characterization of
mathematical knowledge, premise selection, ATPs that use machine
learning, feedback loops between proving and learning, and some
related symbolic classification problems.

Recently, (large) language models (LMs/LLMs) have shown the ability to
generate (not necessarily correct) informal math texts. Beyond
high-school tasks or mathematics olympics tasks (solutions for which
are abundant on the web), their ATP performance in fair-resource
evaluations has, however, so far been questionable %
compared to
targeted architectures such as signature-independent graph neural
networks~\cite{graph2tac}.  Other research questions currently debated in the context of
LLMs are, e.g., the (lack of) \emph{emergence}~\cite{NEURIPS2023_adc98a26}
and the \emph{memorization} of all the benchmark problems, proofs and their informal
presentations available on the web~\cite{JohanssonS23}.\footnote{In particular, while an exact formal
  proof of a solved problem $P$ may or may not be available on the web or
  GitHub (and thus in the LLM training data), it is still quite likely
  that an informal proof with the essential proof ideas for $P$ has
  been seen during the LLM training. In such cases, the LLM is more likely
  doing translation (autoformalization) rather than the usual proving of previously
  unseen problems. Similar issues appear with methods such as (pre-)training on ``synthetic'' problems, which may be generated in ways that make them close to the target ``unseen'' problems.} Perhaps the most promising uses of (not
necessarily large) language models today seem to be the conjecturing
tasks mentioned in Section \ref{nnm} (e.g.,
\cite{UrbanJ20,Rabe0BS21,GauthierOU23,JohanssonS23}), and in autoformalization
\cite{KaliszykUVG14,KaliszykUV17,WangKU18,jiang2023draft}.

On the other hand, systems like \emph{ChatGPT} have recently convinced
a lot of lay and expert audience about the potential of AI systems
trained over a lot of informal knowledge. It is likely that this will
lead to increased efforts in training AI/TP systems, which not only
absorb a lot of knowledge, but also learn to use it correctly and are
capable of self-improvement and new discoveries without hallucination,
thanks to the ground logical layer.
The brief
history of the AI/TP field so far demonstrates that perhaps the most
interesting systems and research directions emerge when the deductive,
search and symbolic methods are in nontrivial ways combined with the
learning, inductive and statistical methods, leading to complex and
novel AI architectures. In this sense, the future of AI/TP research seems to be bright and very open.

\section*{Acknowledgments} 
We thank the whole AITP
community involved in starting and developing these topics, and especially our
AI/TP collaborators and colleagues (often from Prague, Innsbruck, and Nijmegen), including Bob
Veroff, Stephan Schulz, Konstantin Korovin, Herman Geuvers, Tom Heskes,
Sean Holden, Jasmin Blanchette, Chad
Brown, Mikol\'{a}\v{s} Janota, Karel Chvalovsk\'y, Ji\v{r}\'i Vysko\v{c}il, Petr Pudl\'ak, Petr \v{S}t\v{e}p\'anek, Mirek Ol\v{s}\'ak,
Zsolt Zombori, Geoff Sutcliffe, Christian Szegedy, Tom Hales, Larry Paulson, Daniel K\"{u}hlwein, Twan van Laarhoven, Evgeni
Tsivtsivadze, Jelle Piepenbrock, Jan Heemstra, Freek Wiedijk, Robbert Krebbers, Henk Barendregt,
Adam Pease, Moa Johansson, Sarah Winkler, Sara Loos, Ramana Kumar,
Michael Douglas, Michael Rawson, Mario Carneiro, Bartosz Piotrowski, Yutaka Nagashima, Zar Goertzel,
Michael Faerber, Shawn Wang, Jan H\r{u}la, Filip B\'artek and Liao Zhang.

This work has been partially supported by the COST Action CA20111 (CK,
DC, MS), ERC PoC grant no.~101156734 \emph{FormalWeb3} (CK, JJ),  project RICAIP no.~857306
under the EU-H2020 programme (MS), project CORESENSE no.~101070254
under the Horizon Europe programme (MS), ERC-CZ grant no. LL1902
POSTMAN (JJ, TG, JU), Amazon Research Awards (LB, TG, JU), the EU
ICT-48 2020 project TAILOR no.~952215 (LB, JU), the ELISE EU ICT-48
project's (no.~951847) Mobility Program (LB), and the Czech Science
Foundation project 24-12759S (MS).

\bibliographystyle{abbrv} 
\bibliography{cas-refs}

\end{document}

%% file: thol.tex
\makeatletter

\def\holdocspecials{\do\ \do\$\do\&%
  \do\#\do\^\do\^^K\do\_\do\^^A\do\%}

\def\holtt{\trivlist \item[]\if@minipage\else\vskip\parskip\fi
\leftskip\@totalleftmargin\rightskip\z@
\parindent\z@\parfillskip\@flushglue\parskip\z@
\@tempswafalse \def\par{\if@tempswa\hbox{}\fi\@tempswatrue\@@par}
\obeylines \tt \let\do\@makeother \holdocspecials
 \frenchspacing\@vobeyspaces}

\makeatother

\newlength{\hsbw}
\newcommand\HOLSpacing{13pt}

\newenvironment{holnb}{\begin{flushleft}
 \begin{minipage}[b]{\hsbw}
 \vspace*{.06in}
 \begingroup\small\baselineskip\HOLSpacing\footnotesize
 \begin{holtt}}{\end{holtt}\endgroup
 \end{minipage}
 \end{flushleft}}

   \newcommand\hilbert{\varepsilon}
   
   \newcommand{\Thm}{\(\vdash\)}
   \newcommand{\Cond}{\(\rightarrow\)}
   \newcommand{\Eqv}{\(\equiv\)}
   \newcommand{\Iff}{\(\Longleftrightarrow\)\hspace{-1.5mm}}
   \newcommand{\Fa}{\(\forall\)}
   \newcommand{\Et}{\(\exists\)}
   \newcommand{\Eu}{\(\exists_{unique}\)}

   \newcommand{\Impl}{\(\Longrightarrow\)\hspace{-1.5mm}}
   \newcommand{\Func}{\(\to\)\hspace{-1.5mm}}

   \newcommand{\Lam}{\(\lambda\)}
   
   \newcommand{\Minus}{\(-\)}
   \newcommand{\Lminus}{\(-\)\hspace{-1.5mm}}
   \newcommand{\Prime}{\('\)}
   \newcommand{\Und}{\_}
   \newcommand{\Lt}{\(<\)}
   \newcommand{\Gt}{\(>\)}
   \newcommand{\Leq}{\(\leq\)}
   \newcommand{\Geq}{\(\geq\)}
   \newcommand{\Eq}{\(=\)}
   \newcommand{\Lrb}{\((\)}
   \newcommand{\Rrb}{\()\)}

   \newcommand{\Next}{\(\bigcirc\)}
   \newcommand{\Prev}{\(\ominus\)}
   \newcommand{\WPrev}{\(\widetilde{\bigcirc}\)}
   \newcommand{\Event}{\(\Diamond\)}
   \newcommand{\Once}{\(\underline{\Diamond}\)}  
\newcommand{\Hilbert}{\(\hilbert\)}

\newcommand{\Conj}{\(\wedge\)}
\newcommand{\Disj}{\(\vee\)}
\newcommand{\Neg}{\(\neg\)}
\newcommand{\Pnd}{\(\Diamond\)}

\newcommand{\Models}{\(\models\)}

\long\def\holthm#1{{\def\Turns{\Thm} \rechol#1\end\end\end}}

\long\def\rechol#1#2#3{\let\next=\rechol\def\postnext{#2#3}\ifx#1\end
\let\next=\relax\def\postnext{\relax}
\else\ifx#1!\Fa                                          
\else\ifx#1@\Hilbert                                     
\else\ifx#1\#\Pnd                                        
\else\ifx#1'\Prime                                       
\else\ifx#1~\Neg                                         
\else\ifx#1\~\Neg
\else\ifx#1_\Und                                         
\else\ifx#1(\ifx#2+\ifx#3)\Next\def\postnext{}\fi        
            \else\ifx#2-\Prev\def\postnext{}             
            \else\ifx#2~\ifx#3)\WPrev\def\postnext{}\fi            
             \else\Lrb\fi\fi\fi                          
\else\ifx#1)\Rrb%
\else\ifx#1\/\Disj                                       
\else\ifx#1\.\Lam                                        
\else\ifx#1>\ifx#2=\Geq\def\postnext{#3}\else\Gt\fi      
\else\ifx#1?\ifx#2!\Eu\def\postnext{#3}\else\Et\fi       
\else\ifx#1-\ifx#2>\Func\def\postnext{#3}               
            \else\ifx#2-\Lminus\def\postnext{#3}
            \else\Minus\fi\fi                               
\else\ifx#1|\ifx#2-\Turns\def\postnext{#3}               
            \else\ifx#2=\Models\def\postnext{#3}
                 \else\Bar\fi\fi
\else\ifx#1<\ifx#2=\ifx#3>\Iff\def\postnext{}       
                   \else\Leq\def\postnext{#3}\fi    
            \else\ifx#2+\Event\def\postnext{}       
            \else\ifx#2-\Once\def\postnext{}       
            \else\Lt\fi\fi\fi                       
\else\ifx#1=\ifx#2=\ifx#3>\Impl\def\postnext{}            
                   \else\Eqv\def\postnext{#3}\fi         
            \else\ifx#2>\Cond\def\postnext{#3}
                 \else\Eq\fi\fi
\else\ifx#1/\ifx#2\^^M\Conj\par\def\postnext{#3}         
            \else\ifx#2\ \Conj\ \def\postnext{#3}\else#1\fi\fi  
\else#1\fi\fi\fi\fi\fi\fi\fi\fi\fi\fi\fi\fi\fi\fi\fi\fi\fi\fi\fi
\expandafter\next\postnext}